\documentclass[
twocolumn,
]{ceurart}

\usepackage{listings}
\usepackage[linesnumbered,ruled,vlined]{algorithm2e}
\SetKwInput{KwInput}{Input}                
\SetKwInput{KwOutput}{Output}              
\SetKwInput{KwParameters}{Parameters}              

\usepackage{graphicx}
\usepackage{caption}
\usepackage{subcaption}

\begin{document}

\copyrightyear{2022}
\copyrightclause{Copyright for this paper by its authors. Use permitted under Creative Commons License Attribution 4.0 International (CC BY 4.0).}

\conference{The IJCAI-ECAI-22 Workshop on Artificial Intelligence Safety (AISafety 2022), July 24-25, 2022, Vienna, Austria}

\title{Utilizing Class Separation Distance for the Evaluation of Corruption Robustness of Machine Learning Classifiers}

\author[1,2]{Georg Siedel}[%
email=siedel.georg@baua.bund.de,
url=https://github.com/georgsiedel/minimal-separation-corruption-robustness,
]
\cormark[1]
\address[1]{Federal Institute for Occupational Safety and Health (BAuA) Germany}
\address[2]{University of Stuttgart, Germany}

\author[1]{Silvia Vock}
\author[2]{Andrey Morozov}
\author[1]{Stefan Voß}

\cortext[1]{Corresponding author.}

\begin{abstract}
Robustness is a fundamental pillar of Machine Learning (ML) classifiers, substantially determining their reliability. Methods for assessing classifier robustness are therefore essential. In this work, we address the challenge of evaluating corruption robustness in a way that allows comparability and interpretability on a given dataset. We propose a test data augmentation method that uses a robustness distance $\epsilon$ derived from the datasets minimal class separation distance. The resulting MSCR (minimal separation corruption robustness) metric allows a dataset-specific comparison of different classifiers with respect to their corruption robustness. The MSCR value is interpretable, as it represents the classifiers avoidable loss of accuracy due to statistical corruptions. On 2D and image data, we show that the metric reflects different levels of classifier robustness. Furthermore, we observe unexpected optima in classifiers robust accuracy through training and testing classifiers with different levels of noise. While researchers have frequently reported on a significant tradeoff on accuracy when training robust models, we strengthen the view that a tradeoff between accuracy and corruption robustness is not inherent. Our results indicate that robustness training through simple data augmentation can already slightly improve accuracy.
\end{abstract}


\maketitle

\section{Introduction}

ML functions are deployed to an increasing extent over various industries including machinery engineering. Within the European domestic market, machinery products are subject to regulation of the Machinery directive, which demands a risk assessment\footnote
{Machinery Directive, Directive 2006/42/EC of the European Parliament and of the Council of 17 May 2006.}.

Risk assessment includes risk estimation and evaluation, where risk is defined as a combination of probability and severity of a hazardous event. Therefore, once ML functions are deployed in machinery products, where their failure may lead to a hazardous event, being able to quantify the probability and severity of their failures becomes mandatory. However, there still exists a gap between the regulative and normative requirements for safety critical software and the existing methods to assess ML safety \cite{Siedel.2021}.

This work targets ML classifiers, the failures of which are misclassifications. Our focus is on the evaluation of failure probability specifically, not on failure severity. We address one specific failure mode of ML classifiers: Corrupted or perturbed data inputs that cause a change of the output to a misclassification. The property of a classifier resistant to any such input corruptions is called robustness\footnote
{Robustness includes resistance to any corruption-caused class change, which may not be a failure mode when the original point was already misclassified (cf. footnote 4).}.
A classifier is a function that assigns a class to any $d$-dimensional input $x\epsilon R^d$. Classifier $g$ is robust at a point $x$ within a distance $\epsilon>0$, if $g(x)=g(x')$ holds for all perturbed points $x'$ that satisfy $dist(x-x')\leq\epsilon$ \cite{Weng.2018, Yang.2020}. The $dist$-function can e.g. be an $L_p$-norm distance, while $\epsilon$ can be defined based on physical observations of e.g. which perturbations are imperceptible for humans.

Robustness is considered a desirable property since intuitively, a slightly perturbed input (e.g. an imperceptibly changed image) should not lead to a classifier changing its corresponding prediction. In essence, a robustness requirement demands that within a certain input parameter space around $x$, all points $x'$ have to share the same class. This way, a robustness requirement adds additional information on how the classifier should behave near ground truth data points. Authors therefore argue the importance of robustness, being a fundamental pillar of reliability \cite{Zhao.2021} and quality \cite{DeutschesInstitutfurNormung.2020} of ML models. 

However, popular robustness training methods show significantly lowered test accuracy compared to standard training, which has lead to some authors discussing an inherent, i.e. inevitable tradeoff between accuracy and robustness (see Section 2.2). 

Two types of robustness need to be clearly distinguished \cite{DeutschesInstitutfurNormung.2020, Fawzi.2018, Gilmer.2019}: adversarial robustness and corruption robustness.

Adversarial inputs are perturbed data deliberately optimized to fool a classifier into changing its output class. Corruption robustness (sometimes: statistical robustness) describes a model’s output stability not against such worst-case, but against statistically distributed input corruptions. The two types of robustness require different training methods and are differently hard to achieve depending on the data dimension \cite{Fawzi.2018}. In practice, training a model for one of the two robustness types only shows limited or selective improvement for the other type \cite{Gilmer.2019, Hendrycks.2019, Rusak.2020}.

In the field of research towards ML robustness, most of the attention has been given to adversarial attack and defense methods. However, from the perspective of machinery safety and risk assessment, adversarial robustness is mainly a security concern and therefore not in the scope of this article. \cite{Wang.2021} argue that instead of adversarial robustness evaluation, a corruption robustness evaluation is often more applicable to obtain a real-world robustness measure and it can be used to estimate a probability of failure on potentially perturbed inputs for the overall system. 

Contribution: In this paper, we investigate corruption robustness using data augmentation for testing and training\footnote
{Code available on Github, see front page.}.
Our key contributions are twofold:
\begin{itemize}
\item We propose the $"MSCR"$ metric to evaluate and compare classifiers corruption robustness. The approach is independent of prior knowledge about corruption distances, but utilizes properties of the underlying dataset, giving the metric a distinct interpretable meaning. We show experimentally, that the metric captures different levels of classifier corruption robustness.
\item We evaluate the tradeoff between accuracy and robustness from the perspective of corruption robustness and present arguments against the tradeoff being inherent. 
\end{itemize}
After giving an overview of related work, we present our approach for the MSCR metric in section 3.1. We then test our approach on simple 2D as well as image data with the setup described in section 3.2. We present and discuss the results in sections 4 and 5.

\section{Related Work}
\subsection{Measuring corruption robustness}

Corruption robustness of classifiers can be numerically evaluated by testing the ratio of correctly/incorrectly classified inputs from a corrupted test dataset. This ratio is called robust accuracy/error, in contrast to the ratio of correct classification on original test data (“clean accuracy/error”). Robust accuracy represents a combined measure for accuracy and robustness\footnote
{The term astuteness can be used for robust accuracy to differentiate the term from robustness, see \cite{Yang.2020}. Throughout this work, we use the popular term robustness to describe our metric for consistency with works like \cite{Hendrycks.2019} and \cite{Wang.2021}.}. 
A useful way to obtain a measure of robustness only is by subtracting robust accuracy/error and clean accuracy/error \cite{Hendrycks.2019, Lopes.2019}.

In most cases, the corrupted test dataset is derived from an original test dataset through data augmentation. One or multiple corruptions out of some distribution are added to every original data points. Figure 1 explains this procedure of data augmentation with corruptions (dots) being added to a test dataset (stars) with 2 parameters and 2 classes. 
\begin{figure}
  \centering
  \includegraphics[width=\linewidth]{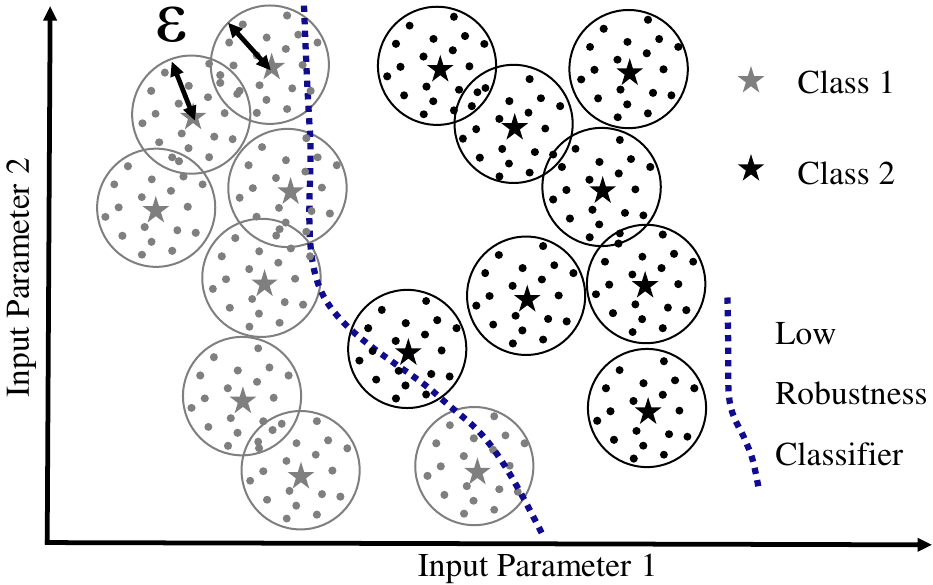}
  \caption{A robustness requirement (here: $L_2$-norm balls with maximum distance $\epsilon$) assigned to the data points (stars) of a 2D binary dataset (2 input parameters, 2 classes). The shown classifier is not robust, since its dotted decision boundary violates the robustness requirement. To evaluate this, additional points (dots) are augmented within $\epsilon$ of each original point. On those points, the robust accuracy of the classifier is measured – for this classifier, some errors arise.}
  \vspace{-5pt}
\end{figure}
It illustrates how a 100\% accurate but non-robust classifier achieves lower robust accuracy on the augmented data points.

The corruption distribution can be defined e.g. based on physical observations. For the example of image data, \cite{Hendrycks.2019, Paterson.2021} add corruptions like brightness, blur and contrast, while \cite{Molokovich.2021, Schwerdtner.2020} use special weather or sensor corruptions. \cite{Hendrycks.2019} created robustness benchmarks for the most popular image datasets based on such physical corruptions.

Corruption distributions can also be defined without physical representations by adding e.g. Gaussian-, salt-and-pepper-, or uniformly distributed noise of certain magnitude to the inputs \cite{Hendrycks.2019, Lopes.2019, Schwerdtner.2020, Wang.2021}. Figure 1 exemplary demonstrates uniformly distributed noise within $L_2$-norm distance $\epsilon$ (in 2D, $L_2$-norm is a circle) of the data points.

With PROVEN, \cite{Weng.2019} propose a framework that uses statistical data augmentation to estimate bounds on adversarial robustness of a model, essentially combining the evaluation of both adversarial and corruption robustness. 

\cite{Zhao.2021} take a robustness evaluation approach different from measuring robust accuracy. The authors augment the entire input space with uniformly distributed data points, independent of a test dataset. They divide the input space into cells, the size of which is based on the r-separation distance described in \cite{Yang.2020} and in section 2.2. This way, they can assign a conflict free ground truth class to each cell and evaluate the misclassification ratio on all added data points. The approach allows for statistical testing of the entire input space, but does not scale well to high dimensions.

An analytical way of measuring the robustness of a classifier is through describing the characteristics of its decision boundary. One possibility is to estimate the local Lipschitzness, i.e. a tightened continuity property of models in proximity to data points. To the best of our knowledge however, Lipschitzness has only been used to investigate adversarial, not corruption robustness \cite{ Weng.2018, Yang.2020}.

Both the measure in \cite{Zhao.2021} and Lipschitzness values lack distinct interpretability in terms of what the calculated value represents exactly.

\subsection{The Accuracy-Robustness-Tradeoff}

Significant effort has recently been put into increasing classifier robustness, commonly targeting adversarial robustness, e.g. in \cite{Rusak.2020, Carmon.2019, Cohen.2019, Madry.2018, Zhang.2019}. All these methods cause a significant drop in clean accuracy.

\cite{Lopes.2019, Hendrycks.2020} and \cite{Wang.2021} observe a clear tradeoff between corruption robustness and accuracy for different training methods using data augmentation. The two former works then propose specialized training methods for mitigating parts of this tradeoff on the popular image datasets CIFAR-10 and ImageNet.

Based on such research, \cite{Zhang.2019} and \cite{Raghunathan.2020} discuss a tradeoff between accuracy and robustness, while \cite{Tsipras.2019} even argue that the cause for this tradeoff is inherent, i.e. inevitable. A counterargument is presented by \cite{Yang.2020}, who argue that accuracy and robustness are not necessarily at odds as long as data points from different classes are separated far enough from each other (see section 3). The authors measure this “r-separation” between different classes on various image datasets and find it to be high enough for classifiers to be both accurate and robust for typical perturbation distances.

\section{Method}

Our robustness evaluation approach is based on this same idea by \cite{Yang.2020}, who measure the distance 2r for a dataset, which is the minimal distance between any two points of different classes (2r in Figure 2).
\begin{figure}
  \centering
  \includegraphics[width=\linewidth]{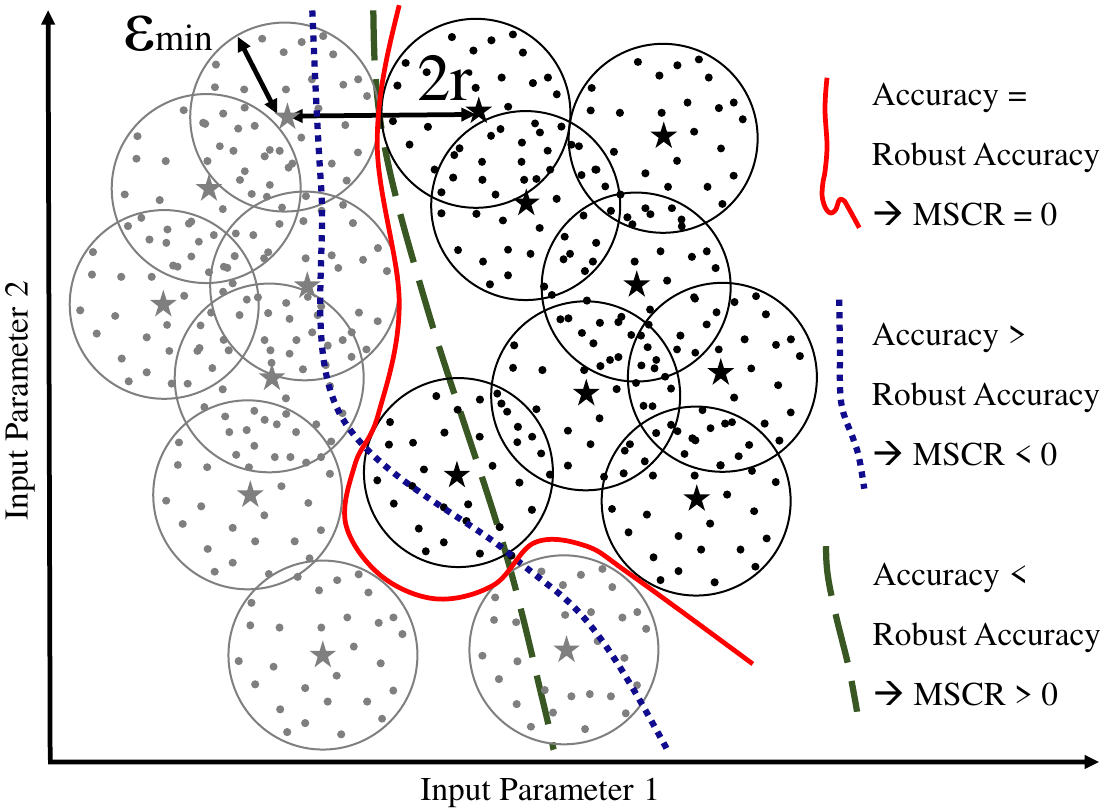}
  \caption{The MSCR concept, demonstrated on 2D test data. Data augmentation is carried out like in Figure 1. The distance ($\epsilon_{min}$) is determined by the minimal distance ($2r$) of original points from different classes (black and grey). This way, augmented points of different classes are still separated and classifiers can be both accurate and robust. The decision boundaries of 3 hypothetical classifiers are shown to demonstrate different levels of robustness and their resulting MSCR value.}
  \vspace{-5pt}
\end{figure}
The authors argue that a classifier can be both robust and accurate as long as
\begin{equation}
  \epsilon\leq r
\end{equation}
holds, where $\epsilon$ is the corruption distance for which robustness is evaluated and $r$ is half this minimal class separation distance. We adopt this notation and set $\epsilon_{min}=r$ as our corner case corruption distance (see Figure 2). The value $\epsilon_{min}$ is not related to any prior physical knowledge of e.g. which corruptions are imperceptible, but is specific for the given dataset, i.e. it is based on the fundamental property of minimal class separation. Accordingly, we call our metric “Minimal Separation Corruption Robustness” (MSCR).

\subsection{MSCR metric}

To measure corruption robustness, we carry out data augmentation on the test data with uniformly distributed corruptions, generated by a random sampling algorithm, similar to the method shown by \cite{Wang.2021}. In contrast to \cite{Wang.2021}, we set the upper bound  of the distance $\epsilon_{test}$, within which the augmented noise is distributed, to $\epsilon_{min}$, as required in Equation 1 (see Figure 2 for an illustration). We measure robust accuracy on the augmented data, which corresponds to a combination of clean accuracy and corruption robustness. However, we want to quantify robustness independent of clean accuracy for comparability, so we subtract the clean accuracy ($Acc_{clean}$) from the robust accuracy on $\epsilon_{min}$-augmented test data ($Acc_{rob-\epsilon_{min}}$) and normalize by the clean accuracy:
\begin{equation}
    MSCR=(Acc_{rob-\epsilon_{min}}-Acc_{clean})/Acc_{clean}
\end{equation}
According to \cite{Yang.2020}, a classifier can in principle be robust on such augmented noise of magnitude $\epsilon_{min}$ while maintaining accuracy. This can be seen from Figure 2, where the circles of radius $\epsilon_{min}$ in which data is augmented, never overlap for different classes. We use an identical radius $\epsilon_{min}$ for all classes, assuming that the separation of data points from the classifiers decision boundary is equally important for all classes. For this noise level $\epsilon_{min}$, any non-robust behavior is theoretically avoidable, since a classifiers decision boundary can separate the classes even with augmented data, as long as the ML algorithm is capable of learning the exact function. The MSCR metric therefore measures the (relative) win or loss in accuracy when testing on such noisy data that any loss is just about avoidable. Figure 2 illustrates the impact of the proposed metric using three corner cases:
\begin{itemize}
\item 	$MSCR=0$, $Acc_{rob-\epsilon_{min}}=Acc_{clean}$, solid line in Figure 2: A classifier that is as robust as possible for the given class separation of the dataset. It not only correctly classifies the original data points, but also all augmented data points.
\item 	$MSCR<0$, $Acc_{rob-\epsilon_{min}}<Acc_{clean}$, dotted line in Figure 2: A classifier that is not perfectly robust. It correctly classifies all original data points, but misclassifies a number of augmented data points due to low robustness.
\item 	$MSCR>0$, $Acc_{rob-\epsilon_{min}}>Acc_{clean}$, dashed line in Figure 2: A classifier misclassifies some original data points, but correctly classifies some of their augmentations. Especially for classifiers that trained to be very robust, we expect this result to be possible.
\end{itemize}
Algorithm 1 shows the MSCR calculation procedure. In step 1, different distance functions (e.g. $L_\infty$-norm) can be applied. We account for randomness in the data splitting, model training and data augmentation procedures by carrying out multiple runs of the same experiment and reporting average values and 95\%-confidence intervals over all runs. The reasonable number of augmented points k per original data point varies depending on the dataset (see section 3.2). Within the respective for-loop, variable $models$ runs through the list of all classifier models to be compared, while $r$ counts up to (the overall number of) $runs$.

\begin{algorithm}[!ht]
\DontPrintSemicolon
  \KwData{classification dataset $\{X(x_1,…,x_n), Y(y_1,…,y_n)\}$}
  \KwParameters{$models =\{model_1,…,model_m\}$, $r=\{1,…,runs\}$, $k$, $\epsilon_{test}=\{0, \epsilon_{min}\}$}
  \KwOutput{$\overline{MSCR}=\{\overline{MSCR_1},…,\overline{MSCR_m}$\}}

  $\epsilon_{min}=(\min\limits_{x_{i \epsilon n}, x_{j \epsilon n}}\{dist(x_j-x_i)|y_i\neq y_j \})/2$ \;
  \For{$models$}
    {
        \For{$r$}    
        { 
        	Train $model_m$\;
        	Test model with original test data ($\epsilon_{test}=0$) $\rightarrow$ 
			return $Acc_{clean}$\;						
            For every test data point: Uniform random sample $k$ points within $dist(\epsilon_{min})$ and augment the test data\;
            Test model with data from step 6 $\rightarrow$ return $Acc_{rob-\epsilon_{min}}$\;
            $MSCR_{r}=(Acc_{rob-\epsilon_{min}}-Acc_{clean})/Acc_{clean}$\;
        }
        $\overline{MSCR_m}=(\sum\nolimits_{r=1}^{runs} MSCR_{r})/runs$\;
    }
\caption{MSCR calculation}
\vspace{-2pt}
\end{algorithm}

\subsection{Experimental details}

Additionally to test data augmentation, we train multiple models on datasets augmented with different corruption distances $\epsilon_{train}$. Increasing a model’s $\epsilon_{train}$ should lead to a growing MSCR value, as it is expected that the model robustness grows. This way, we evaluate the trend of the MSCR value for models with different corruption robustness levels. Also, on test data corrupted with large $\epsilon_{test}$, models trained with $\epsilon_{train}=\epsilon_{test}$ are expected to perform best \cite{Wang.2021}. 

As demonstrated in Figure 2, corruption levels below $\epsilon_{min}$ theoretically allow a classifier to be robust while not losing test accuracy. We investigate this theoretical claim by \cite{Yang.2020} additionally to the MSCR metric by evaluating changes in robust accuracy when augmenting multiple corruption levels $\epsilon_{test}$ to the test dataset. In contrast to the work of \cite{Wang.2021}, we extensively evaluate more corruption levels below, around and including $\epsilon_{min}$ specifically. In contrast to the work of \cite{Lopes.2019} and \cite{Hendrycks.2020}, we use simple uniformly distributed data augmentation with a fixed upper bound of noise for the entire dataset instead of Gaussian noise. This allows us the comparison of the noise levels with the class separation distances. It shall be noted however that in contrast to Gaussian noise, where density decreases with distance, uniform noise does not reflect the higher uncertainty in a class assignment when the distance from a ground truth data point increases. Even though our data augmentation method is simple, we still expect to find counterexamples for the accuracy-robustness-tradeoff, based solely on the class-separation theory. We believe that the case of finding such counterexamples with less advanced methods than e.g. \cite{Lopes.2019} represents even more credible evidence for the argument of \cite{Yang.2020} against an inherent accuracy-robustness-tradeoff.

We carry out the experiments on 3 binary class 2D datasets as were used and provided by \cite{Zhao.2021}. For clarity, we only report results with $L_\infty$-corruptions on one of those datasets, which is shown in Figure 3 and features 4674 data points. 
\begin{figure}
  \centering
  \includegraphics[width=165pt]{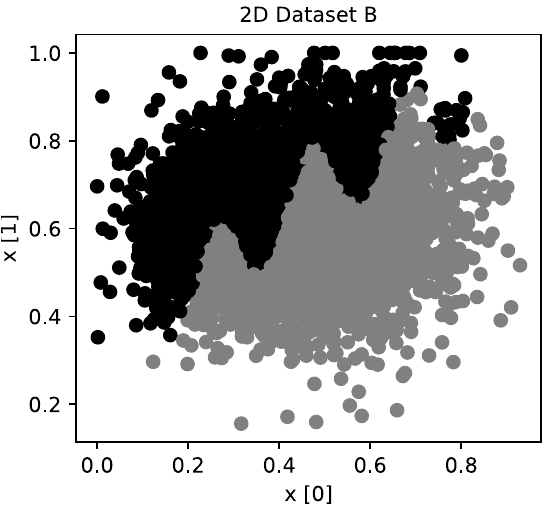}
  \vspace{-5pt}
  \caption{Data points in the binary class 2D dataset.}
  \vspace{-8pt}
\end{figure}
Experiments with the other 2D datasets and $L_2$-corruptions exhibit similar fundamental results, which can also be found in our Github repository (see frontpage). The two input parameters $x[0]$ and $x[1]$ are normalized to the interval $[0, 1]$. For classification, we use a random forest (RF) algorithm with 100 trees. We also compare this classifier with a 1-nearest-neighbor model, which is known to be inherently robust, since it classifies based on distance to the 1 nearest data point. 

We choose $k=10$ augmented data points per original data point, as we found higher numbers of $k$ not significantly improving the resulting robust accuracy and its standard deviation. This effect of different values for the hyperparameter $k$ is displayed in Figure 4. In order to achieve statistically representative results, we evaluate how the average test accuracy converges over multiple runs and accordingly choose 1200 runs.

The experiments are additionally run in a more applied image classification setting using benchmark dataset CIFAR-10. We adopt the classifier architecture from \cite{Wang.2021}, using a 28-10 wide residual network with SGD optimizer, 0.3 dropout rate, training batch size 32 and 30 epochs with a 3-step decreasing learning rate. All pixel values are normalized to $[0, 1]$ and random horizontal flips and random crops with 4px padding are used for training generalization. For CIFAR-10 we choose $k=1$, since \cite{Wang.2021} report one augmented point to be sufficient. We suspect that this is due to the multiple epochs of the training process, which allows to train the model on multiple augmentations per training data point. We choose 20 runs due to computational feasibility of all training procedures.
\begin{figure}
  \centering
  \includegraphics[width=\linewidth]{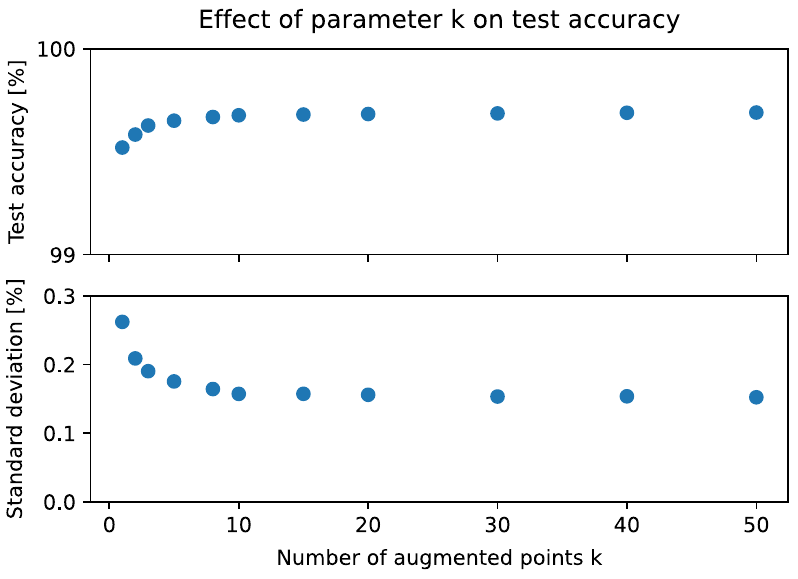}
    \vspace{-12pt}
  \caption{Effect of hyperparameter $k$ on robust accuracy and its deviation. 2D dataset, $\epsilon_{train},\epsilon_{test}=0.001$.}
  \vspace{-3pt}
\end{figure}
Table 1 shows the minimal class separation distances $2r$ and the corresponding $\epsilon_{min}$ values, measured in $L_\infty$-distance for both datasets. For intuition, the CIFAR-10 $\epsilon_{min}$ value translates to a maximum color grade change of $27/255$ on all pixels. Higher values for $2r$ are to be expected for image data, since $L_\infty$-norm evaluates the maximum distance in any of the 3072 dimensions of CIFAR-10 input data.
\begin{table}
  \caption{Minimal $L_\infty$ class separation and corresponding $\epsilon_{min}$}
  \label{tab:separation}
  \begin{tabular}{ccl}
    \toprule
    Dataset & $2r(L_\infty)$ & $\epsilon_{min}$\\
    \midrule
    2D dataset & 0.008026 & 0.004013 \\
    CIFAR-10 (train and test set) & 0.211765 & 0.105882 \\
    \bottomrule
  \end{tabular}
\end{table}

\section{Results}

Table 2 displays the matrix of test accuracies for the 2D dataset for different values of both $\epsilon_{train}$ (representing different models, along columns) and $\epsilon_{test}$ (along rows). The bold values highlight the best model for every level of test noise. As can be seen, the optima of the accuracy do not actually match with the matrix diagonal, where training and test noise are equal (highlighted in light grey). Instead, when testing with lower noise levels and even with clean test data, the model trained on $\epsilon_{train}=0.007$ performs best. The maximum overall accuracy is achieved with a model trained on $\epsilon_{train}=0.007$ that is also tested on $\epsilon_{test}=0.001$ corruptions. For higher noise levels, the optimum robust accuracies are achieved with $\epsilon_{train}\leq\epsilon_{test}$, displaying the opposite trend compared to low noise levels.

The results on CIFAR-10 in Table 3 show a similar trend, albeit less pronounced. For low noise levels, training with $\epsilon_{train}=0.01$ appears to be optimal for clean accuracy. The maximum overall accuracy is achieved with $\epsilon_{train}=0.02$ and $\epsilon_{test}=0.01$. For higher levels of test noise, similarly to 2D data, it appears beneficial to use $\epsilon_{train}\leq\epsilon_{test}$. In contrast to the 2D data, where the optimum $\epsilon_{train}$ for $\epsilon_{test}=0$ is higher than the $\epsilon_{min}$ value, for CIFAR-10 it is $\sim$10 times lower than $\epsilon_{min}$. The optimum $\epsilon_{train}=0.01$ translates to a $2.5/255$ color grade corruption for every pixel. 

\begin{table*}[tcb]
    \caption{Clean accuracies (first row) and robust accuracies in percentage plus the MSCR value (last row) for various models (columns) ± the 95\% confidence intervals. Models are trained and tested with different levels of $L_\infty$-noise ($\epsilon_{train}$ along columns, $\epsilon_{test}$ along rows). Bold accuracies: Best model accuracy for every noise level. Bold MSCR value: Highest MSCR value, i.e. highest model robustness. Last row color scale: Highlights the constant increase of MSCR with increasing $\epsilon_{train}$. Light grey accuracies: Model trained and tested on the same noise level ($\epsilon_{train}=\epsilon_{test}$). Dark grey accuracies: Maximum overall accuracy.}
    \begin{subtable}[a]{\linewidth}
        \centering
        \label{fig:2d_results} 
        \includegraphics[width=\linewidth]{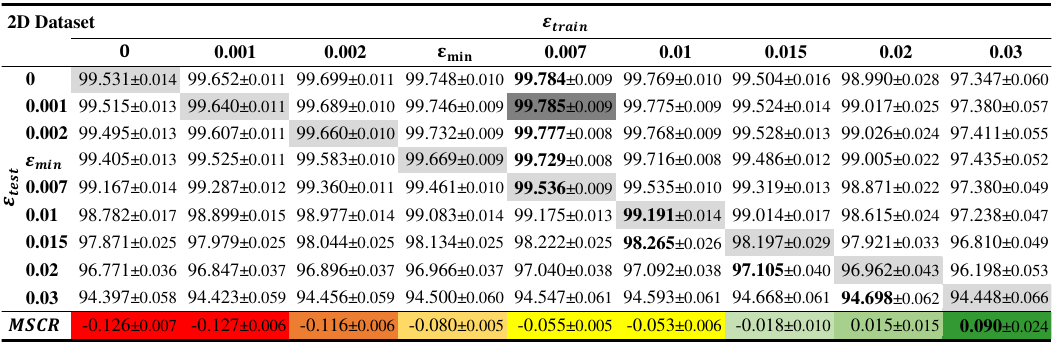}
        \vspace{-13pt}
        \caption{2D Dataset}
        \vspace{+4pt}
    \end{subtable}
    ~
    \begin{subtable}[b]{\linewidth}
        \centering
        \label{fig:cifar_results} 
        \includegraphics[width=\linewidth]{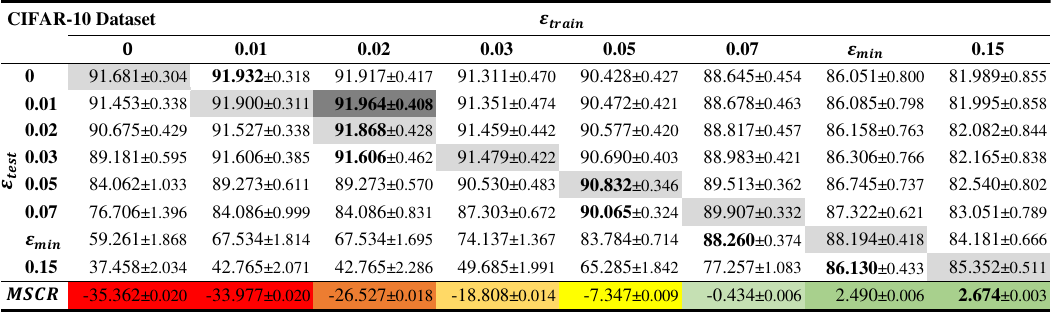}
        \vspace{-13pt}
        \caption{CIFAR-10 Dataset}
    \end{subtable}
    \vspace{-12pt}
    \label{tab:results}
\end{table*}

For both datasets it is visible from the last rows of Table 2 and 3, that the MSCR value steadily increases with higher levels of training noise $\epsilon_{train}$. For both datasets, the MSCR increases from negative values on less robust trained models to zero and even positive values for more robust trained models.

For CIFAR-10, the MSCR values are overall much larger than for the 2D data. This effect correlates with the $\epsilon_{min}$  noise level, which is about 26 times larger in absolute values.

Figure 5 shows a comparison on the 2D dataset between the 1NN model and the RF model with regards to clean accuracy (Fig. 5a) and MSCR (Fig. 5b). Both models are trained on the various $\epsilon_{train}$ values. While for the RF model, both metrics increase with increasing training noise up to the optimum of $\epsilon_{train}=0.007$, the 1NN model shows constant (and superior) metrics up to this training noise. This illustrates the inherent robustness of the 1NN model. The comparison also shows that this inherent robustness is indeed advantageous regarding accuracy on our dataset.

\begin{figure*}[ht]
  \subfloat[Clean Accuracy]{
	\begin{minipage}[c][1\width]{
	   0.49\textwidth}
	   \includegraphics[width=1\textwidth]{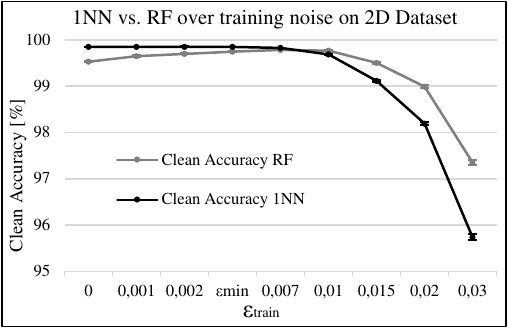}
	\end{minipage}
	\vspace{-38pt}}
 \hfill 	
  \subfloat[Robustness (MSCR)]{
	\begin{minipage}[c][1\width]{
	   0.49\textwidth}
	   \includegraphics[width=1\textwidth]{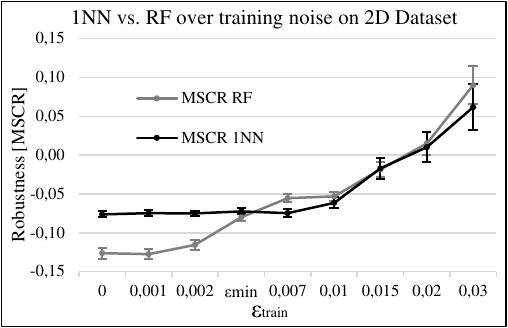}
	\end{minipage}
    \vspace{-38pt}}
\vspace{-3pt}
\caption{Model comparison on 2D Dataset with regards to clean accuracy and robustness (MSCR): RF versus 1NN model with different $\epsilon_{train}$.}
\vspace{-34pt}
\end{figure*}

Figures 6a (2D dataset) and 6b (CIFAR-10) display the accuracy-robustness-tradeoff for the models trained with different $\epsilon_{train}$ by contrasting MSCR versus clean accuracy values. Both Figures in principle show a tradeoff curve. However, it is visible that for $\epsilon_{train}\leq0.007$ on 2D data and $\epsilon_{train}\leq0.01$ on CIFAR-10, both clean accuracy and robustness increase compared to the baseline model with $\epsilon_{train}=0$. The tradeoff is overcome for these models (arguably also for $\epsilon_{train}=0.01$ for 2D data and $\epsilon_{train}=0.02$ for CIFAR-10).

\begin{figure*}[ht]
  \subfloat[2D Dataset]{
	\begin{minipage}[c][1\width]{
	   0.489\textwidth}
	   \includegraphics[width=1\textwidth]{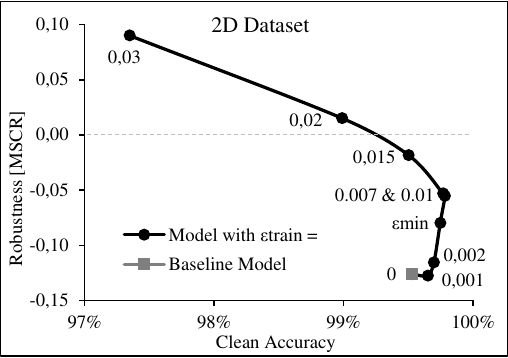}
	\end{minipage}
	\vspace{-32pt}}
 \hfill 	
  \subfloat[CIFAR-10 Dataset]{
	\begin{minipage}[c][1\width]{
	   0.489\textwidth}
	   \includegraphics[width=1\textwidth]{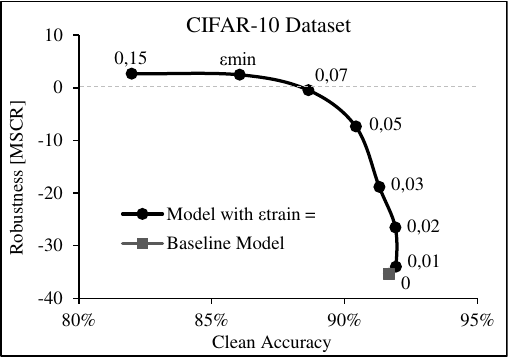}
	\end{minipage}
    \vspace{-32pt}}
\vspace{-3pt}
\caption{Accuracy-robustness-tradeoff for models trained with different levels of augmented training noise $\epsilon_{train}$, compared to the baseline model with $\epsilon_{train}=0$. Models with both higher MSCR and higher clean accuracy (when the curve evolves towards the top right corner) contradict the inherent tradeoff.}
\vspace{-4pt}
\end{figure*}

\section{Discussion}
\subsection{Applicability of the MSCR metric}
Our results from the experiments indicate that the relative difference between the noise-augmented robust accuracy and the clean accuracy is a measure for corruption robustness of models. For $\epsilon_{test}=\epsilon_{min}$ in particular, this relative difference that we named MSCR steadily increases with higher corruption robustness of the RF model on 2D data and the wide residual network on CIFAR-10. This way, we verify the metric’s capability to reflect the corruption robustness of different models. However, this claim is based on the assumption that increasing corruption robustness of our models can be generated through training with higher noise levels. This seems evident based on research by \cite{Wang.2021}, but requires future validation like in \cite{Lopes.2019}, who confirm that their Gaussian robustness metric is strongly correlated with the popular physical corruptions benchmark by \cite{Hendrycks.2019}.

On the 2D dataset, the 1NN model shows a constant, superior MSCR value compared to the RF model for all $\epsilon_{train}\leq0.007$, where classes are still predominantly separated. This is the performance expected from an inherently robust model such as 1NN, which fits its decision boundary based on maximum class separation. The MSCR values are able to correctly display this interrelation.

\subsection{Disadvantages and advantages of the MSCR metric}
In our experiments, the steady robustness increase for higher $\epsilon_{train}$ also holds for other levels of testing noise than $\epsilon_{min}$. The MSCR value, which uses $\epsilon_{min}$-corruptions as the underlying robustness requirement, is only one particular case of this robustness calculation approach. It has to be emphasized that from our results in Tables 2 and 3, we cannot observe any conspicuities for $\epsilon_{test}\sim\epsilon_{min}$. For example, there is no indication that models perform well below this noise level while massively dropping off at higher noise levels, as could be presumed from the r-separation theory. It is therefore evident to conclude that measuring corruption robustness works with other $\epsilon_{test}$-values. In practice, if specific corruptions are known for an application, those corruptions should also be used for testing, e.g. through benchmarks \cite{Hendrycks.2019}. 

However, we emphasize that the MSCR metric is advantageous in two ways: First, it does not require prior physical knowledge to define corruption distributions, like e.g. \cite{Hendrycks.2019} does. Instead, it only requires measuring the actual class separation from any classification dataset. Second, the MSCR can be interpreted with a clear contextual meaning, since the robustness requirement is derived from the dataset: It measures “the theoretically avoidable loss (or win) of accuracy due to statistical corruptions”.

\subsection{On achieving high MSCR values}
Clearly, avoiding any loss of accuracy on $\epsilon_{min}$-noise is hard to achieve in practice on high-dimensional data. For CIFAR-10, $MSCR=0$ can be achieved, but only with $\epsilon_{train}=0.07$, where the clean accuracy declines by 3 percentage points compared to $\epsilon_{train}=0$. We also verify our conjecture that $MSCR>0$ is possible for some robust trained models. For this behavior, we find the discovery in \cite{Mickisch.2020} a convincing technical explanation. Misclassified data points tend to lie closer to the decision boundary than correctly classified data points. The data augmentations on a misclassified data point therefore have a high chance of causing a favorable class change. At the same time, data augmentations on correctly classified points have a lower chance of causing an unfavorable class change when their distance to the decision boundary is high, which is what a robust model is trained for.

\subsection{The accuracy-robustness-tradeoff}
Besides our investigation of the MSCR metric, we report on findings regarding the tradeoff between accuracy and corruption robustness. For both 2D and CIFAR-10 datasets we observe higher clean and robust accuracy on any test noise when training a model with a specific level of uniform noise ($\epsilon_{train}=0.007$ for 2D, $\epsilon_{train}=0.01$ for CIFAR-10), compared to standard training. For the 2D data, this optimum $\epsilon_{train}$ value is even higher than $\epsilon_{min}$, the value which the r-separation theory suggests to be beneficial for robustness while not hurting accuracy. This could be due to the major proportion of minimal distances of data points to other classes being significantly bigger than $\epsilon_{min}$. Our results are statistically significant for the 2D dataset experiment. For 20 runs per trained model on CIFAR-10, we emphasize that claiming higher mean clean accuracy for any $\epsilon_{train}>0$ compared to $\epsilon_{train}=0$ does not achieve 95\%-confidence in a pairwise statistical comparison. More than 20 runs are necessary to obtain statistically significant results, which we could not achieve due to limited computational resources. Hence, we only treat our results on CIFAR-10 regarding the accuracy-robustness-tradeoff as suggestions.

The suggestion that some $\epsilon_{train}>0$ leads to higher clean accuracy than $\epsilon_{train}=0$ has theoretical relevance. It supports the claim made, but not practically proven by \cite{Yang.2020}, that accuracy and robustness are not in an inherent tradeoff as long as the noise level $\epsilon$ fulfills Equation 1. 

The result also seems relevant from a practical perspective, since developers may try some $\epsilon_{train}$ for training data augmentation, which increases robustness without drawbacks regarding accuracy. We emphasize that this practical implication is only valid for the very limited model architectures, datasets and augmentation distributions we tested. For example, our experiments show that noise training below $\epsilon_{min}$ has no effect on an inherently robust model such as 1NN. This is due to the fact that this model type maximizes the class separation of its decision boundary in training anyways.

On the one hand, overcoming the tradeoff for small $\epsilon_{train}$ is not entirely surprising, since it is well known that data transformations and data augmentations can increase generalization of models (in fact, we also used random flips and crops for CIFAR-10 training). \cite{Lopes.2019} and \cite{Hendrycks.2020} also manage to overcome the tradeoff with more advanced training methods. On the other hand, our results are surprising considering this drawback-free increase in robust accuracy is quite significant for the RF model on 2D data (less than halving the classification error). Also, uniform $L_\infty$ data augmentation is a very simple method and less contextually relevant compared to physically derived augmentations. An explanation may be that the uniform $L_p$-norm noise allows a stricter coverage of the input parameter space near data points compared to physical data augmentations, enforcing a smooth model that is less prone to overfitting the corruptions.

\subsection{Class separation distance for model training}
From our results we also need to conclude that in practice, the $\epsilon_{min}$ value has only limited expressiveness when trying to find the optimal $\epsilon_{train}$ with regards to (robust) accuracy. This is visible in Figures 6a and 6b, where based solely on the r-separation theory, we may have expected the curve to reverse its trend along the x-axis when $\epsilon_{train}=\epsilon_{min}$. In reality, the best overall accuracy for the 2D data is achieved for $\epsilon_{train}\sim2*\epsilon_{min}$, while on CIFAR-10 it is achieved for $\epsilon_{train}<\epsilon{_min}/5$. We suspect that high-dimensional datasets are notoriously hard to train with regards to high robust accuracy, at least for such $\epsilon_{min}$ levels their high $L_\infty$ class separation distance inevitably entails. We suspect that on other datasets $\epsilon_{min}$ may be even greater and further away from the optimum $\epsilon_{train}$. Additional research is needed on various distance measures, dataset dimensions and model types in order to utilize class separation distances for optimizing robust accuracy.

\subsection{Optima of $\epsilon_{train}$ vs. $\epsilon_{test}$}
Another interesting finding from the accuracy matrix of both datasets is that the best $\epsilon_{train}$ value for models evaluated with certain  $\epsilon_{test}$ deviates from the expected diagonal. For example, $\epsilon_{train}=0.03$ is not the best choice to prepare for $\epsilon_{test}=0.03$. In Figure 7, the accuracy matrix for CIFAR-10 from Table 3 is visualized in a 3D plot, which shows how the optima in (robust) accuracy deviate from the diagonal. It appears that for low noise levels the best choice is $\epsilon_{train}>\epsilon_{test}$, while for higher noise levels $\epsilon_{train}<\epsilon_{test}$ is more favorable. This suspected dependency needs further investigation. 

\section{Conclusion}
In this article we evaluated a data augmentation method in order to obtain a comparable, interpretable measure of corruption robustness for classifiers. We measured the relative difference between the robust accuracy on corrupted test data and the clean accuracy. We proposed to use half the minimal class separation distance measured from the dataset as the maximum distance $\epsilon_{min}$ of the augmented test noise. This robustness requirement does not presume any prior knowledge about real corruption distances. It theoretically allows a classifier to be fully robust while not losing accuracy. The class separation distance therefore gives our metric a distinct meaning: It represents any “avoidable” loss (or win) in accuracy due to corruptions. We experimentally showed that our metric is able to reflect various degrees of model robustness. 

From training classifiers with different levels of noise we found that classifiers with the highest robust accuracy on a certain level of noise are not strictly those, which are trained on this same level of noise. We also presented indications that a tradeoff between accuracy and corruption robustness is not inherent: In our experiments, simple augmentation training on significant random uniform noise could improve test accuracy of classifiers additionally to their robustness, compared with normal training. However, the minimal class separation distance could in practice not guide us towards the optimal values of training noise. These findings regarding the accuracy-robustness-tradeoff could in our opinion be useful in practice.

Our work seems to fit into a gap between those researchers optimizing test accuracy and those optimizing robustness. Our future work will include further investigations of data augmentation training and testing using other dataset types, distance metrics and corruption distributions. It would be of additional interest, whether some increase in adversarial robustness can be obtained without loosing accuracy. Our findings emphasize the potential and encourage the development of advanced training procedures mitigating the accuracy-robustness-tradeoff, since the combination of both properties is essential from a risk assessment perspective.
\begin{figure}
  \centering
  \includegraphics[width=\linewidth]{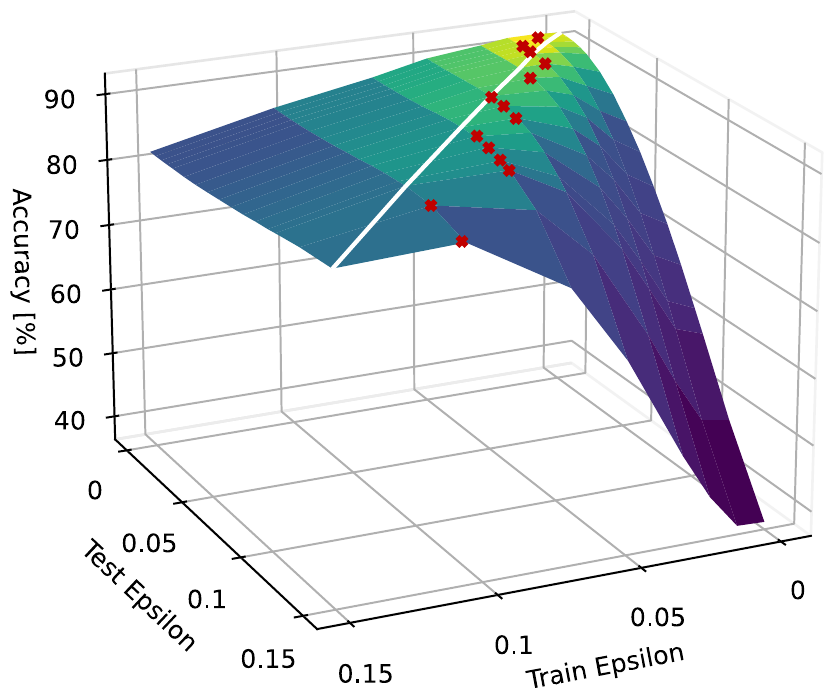}
  \vspace{-4pt}
  \caption{CIFAR-10 (robust) accuracies for different $\epsilon_{train}$ and $\epsilon_{test}$. The optima, marked with points, deviate from the diagonal (white line where $\epsilon_{train}=\epsilon_{test}$): towards higher $\epsilon_{train}$ for lower noise levels and towards lower $\epsilon_{train}$ for higher noise levels.}
\vspace{+3pt}
\end{figure}
\bibliography{refs}

@article{Raghunathan.2020,
 author = {Raghunathan, Aditi and Xie, Sang Michael and Yang, Fanny and Duchi, John and Liang, Percy},
 year = {2020},
 title = {Understanding and mitigating the tradeoff between robustness and accuracy},
 journal = {International Conference on Machine Learning (ICML)}
}

@inproceedings{Rusak.2020,
 author = {Rusak, Evgenia and Schott, Lukas and Zimmermann, Roland S. and Bitterwolf, Julian and Bringmann, Oliver and Bethge, Matthias and Brendel, Wieland},
 title = {A simple way to make neural networks robust against diverse image corruptions},
 pages = {53--69},
 booktitle = {European Conference on Computer Vision},
 year = {2020}
}

@article{Schwerdtner.2020,
 author = {Schwerdtner, Paul and Gre{\ss}ner, Florens and Kapoor, Nikhil and Assion, Felix and Sass, Ren{\'e} and G{\"u}nther, Wiebke and H{\"u}ger, Fabian and Schlicht, Peter},
 year = {2020},
 title = {Risk Assessment for Machine Learning Models},
 url = {https://arxiv.org/pdf/2011.04328.pdf},
 journal = {NeurIPS 2020 Virtual Workshop: Machine Learning for Autonomous Driving}
}

@inproceedings{Siedel.2021,
 author = {Siedel, Georg and Vo{\ss}, Stefan and Vock, Silvia},
 title = {An Overview of the Research Landscape in the Field of Safe Machine Learning},
 publisher = {{American Society of Mechanical Engineers}},
 isbn = {978-0-7918-8569-7},
 booktitle = {Volume 13: Safety Engineering, Risk, and Reliability Analysis; Research Posters},
 year = {2021},
 doi = {10.1115/IMECE2021-69390}}

@article{Tsipras.2019,
 author = {Tsipras, Dimitris and Santurkar, Shibani and Engstrom, Logan and Turner, Alexander and Madry, Aleksander},
 year = {2019},
 title = {Robustness may be at odds with accuracy},
 journal = {International Conference on Learning Representations (ICLR)}
}

@inproceedings{Wang.2021,
 author = {Wang, Benjie and Webb, Stefan and Rainforth, Tom},
 title = {Statistically robust neural network classification},
 pages = {1735--1745},
 booktitle = {Uncertainty in Artificial Intelligence (UAI)},
 year = {2021}
}

@inproceedings{Weng.2019,
 author = {Weng, Lily and Chen, Pin-Yu and Nguyen, Lam and Squillante, Mark and Boopathy, Akhilan and Oseledets, Ivan and Daniel, Luca},
 title = {PROVEN: Verifying robustness of neural networks with a probabilistic approach},
 pages = {6727--6736},
 booktitle = {International Conference on Machine Learning},
 year = {2019}
}

@article{Weng.2018,
 author = {Weng, Tsui-Wei and Zhang, Huan and Chen, Pin-Yu and Yi, Jinfeng and Su, Dong and Gao, Yupeng and Hsieh, Cho-Jui and Daniel, Luca},
 year = {2018},
 title = {Evaluating the robustness of neural networks: An extreme value theory approach},
 journal = {International Conference on Learning Representations (ICLR)}
}

@article{Yang.2020,
 author = {Yang, Yao-Yuan and Rashtchian, Cyrus and Zhang, Hongyang and Salakhutdinov, Russ R. and Chaudhuri, Kamalika},
 year = {2020},
 title = {A closer look at accuracy vs. robustness},
 pages = {8588--8601},
 volume = {33},
 journal = {Advances in neural information processing systems}
}

@inproceedings{Paterson.2021,
 author = {Paterson, Colin and Wu, Haoze and Grese, John and Calinescu, Radu and Pasareanu, Corina S. and Barrett, Clark},
 title = {DeepCert: Verification of Contextually Relevant Robustness for Neural Network Image Classifiers},
 pages = {3--17},
 booktitle = {International Conference on Computer Safety, Reliability, and Security},
 year = {2021}
}

@article{Carmon.2019,
 author = {Carmon, Yair and Raghunathan, Aditi and Schmidt, Ludwig and Duchi, John C. and Liang, Percy S.},
 year = {2019},
 title = {Unlabeled data improves adversarial robustness},
 volume = {32},
 journal = {Advances in neural information processing systems}
}

@inproceedings{Cohen.2019,
 author = {Cohen, Jeremy and Rosenfeld, Elan and Kolter, Zico},
 title = {Certified adversarial robustness via randomized smoothing},
 pages = {1310--1320},
 booktitle = {International Conference on Machine Learning},
 year = {2019}
}

@misc{DeutschesInstitutfurNormung.2020,
 year = {2020},
 title = {DIN SPEC 92001-2: Artificial Intelligence -- Life Cycle Processes and Quality Requirements: Part 2: Robustness},
 volume = {ICS 35.080; 35.240.01},
 number = {DIN SPEC 92001-2},
 author = {{Deutsches Institut f{\"u}r Normung}}
}

@article{Fawzi.2018,
 author = {Fawzi, Alhussein and Fawzi, Omar and Frossard, Pascal},
 year = {2018},
 title = {Analysis of classifiers' robustness to adversarial perturbations},
 pages = {481--508},
 volume = {107},
 number = {3},
 journal = {Machine learning}
}

@inproceedings{Gilmer.2019,
 author = {Gilmer, Justin and Ford, Nicolas and Carlini, Nicholas and Cubuk, Ekin},
 title = {Adversarial examples are a natural consequence of test error in noise},
 pages = {2280--2289},
 booktitle = {International Conference on Machine Learning},
 year = {2019}
}

@article{Hendrycks.2019,
 author = {Hendrycks, Dan and Dietterich, Thomas},
 year = {2019},
 title = {Benchmarking neural network robustness to common corruptions and perturbations},
 journal = {International Conference on Learning Representations (ICLR)}
}

@article{Hendrycks.2020,
 author = {Hendrycks, Dan and Mu, Norman and Cubuk, Ekin D. and Zoph, Barret and Gilmer, Justin and Lakshminarayanan, Balaji},
 year = {2020},
 title = {Augmix: A simple data processing method to improve robustness and uncertainty},
 journal = {International Conference on Learning Representations (ICLR)}
}

@article{Lopes.2019,
 author = {Lopes, Raphael Gontijo and Yin, Dong and Poole, Ben and Gilmer, Justin and Cubuk, Ekin D.},
 year = {2019},
 title = {Improving robustness without sacrificing accuracy with patch gaussian augmentation},
 journal = {arXiv preprint arXiv:1906.02611}
}

@article{Madry.2018,
 author = {Madry, Aleksander and Makelov, Aleksandar and Schmidt, Ludwig and Tsipras, Dimitris and Vladu, Adrian},
 year = {2018},
 title = {Towards deep learning models resistant to adversarial attacks},
 journal = {International Conference on Learning Representations (ICLR)}
}

@article{Mickisch.2020,
 author = {Mickisch, David and Assion, Felix and Gre{\ss}ner, Florens and G{\"u}nther, Wiebke and Motta, Mariele},
 year = {2020},
 title = {Understanding the decision boundary of deep neural networks: An empirical study},
 journal = {arXiv preprint arXiv:2002.01810}
}

@inproceedings{Molokovich.2021,
 author = {Molokovich, O. and Morozov, A. and Yusupova, N. and Janschek, K.},
 title = {Evaluation of graphic data corruptions impact on artificial intelligence applications},
 pages = {012010},
 volume = {1069},
 booktitle = {IOP Conference Series: Materials Science and Engineering},
 year = {2021}
}

@inproceedings{Zhang.2019,
 author = {Zhang, Hongyang and Yu, Yaodong and Jiao, Jiantao and Xing, Eric and {El Ghaoui}, Laurent and Jordan, Michael},
 title = {Theoretically principled trade-off between robustness and accuracy},
 pages = {7472--7482},
 booktitle = {International Conference on Machine Learning},
 year = {2019}
}

@article{Zhao.2021,
 author = {Zhao, Xingyu and Huang, Wei and Bharti, Vibhav and Dong, Yi and Cox, Victoria and Banks, Alec and Wang, Sen and Schewe, Sven and Huang, Xiaowei},
 year = {2021},
 title = {Reliability Assessment and Safety Arguments for Machine Learning Components in Assuring Learning-Enabled Autonomous Systems},
 journal = {arXiv preprint arXiv:2112.00646}
}

\end{document}